%% file: main.tex
\documentclass[lettersize,journal]{IEEEtran}
\usepackage{amsmath,amsfonts}
\usepackage{algorithmic}
\usepackage{algorithm}
\usepackage{array}
\usepackage[caption=false,font=normalsize,labelfont=sf,textfont=sf]{subfig}
\usepackage{textcomp}
\usepackage{stfloats}
\usepackage{url}
\usepackage{verbatim}
\usepackage{graphicx}
\usepackage{cite}
\usepackage{graphicx}
\usepackage{cuted}
\usepackage{caption}
\usepackage{booktabs}
\usepackage{multirow}
\usepackage{adjustbox}
\begin{document}

\title{Casual Flash Lighting for Gaussian Splat Inverse Rendering}


\author{
Jiamin Xu,
Dongheng Wei,
Jiarong Zhao,
Qi Wang,
James Tompkin,
Weiwei Xu,
and Gang Xu%
\thanks{Jiamin Xu, Dongheng Wei, Jiarong Zhao, and Gang Xu are with
Hangzhou Dianzi University, China.
E-mail: superxjm@yeah.net, 241050123@hdu.edu.cn,
252050195@hdu.edu.cn, and gxu@hdu.edu.cn.}%
\thanks{Qi Wang is with North China Electric Power University, China.
E-mail: wqnina1995@gmail.com.}%
\thanks{James Tompkin is with Brown University, USA.
E-mail: james\_tompkin@brown.edu.}%
\thanks{Weiwei Xu is with the State Key Laboratory of CAD\&CG,
Zhejiang University, China.
E-mail: xww@cad.zju.edu.cn.}%
\thanks{Gang Xu is the corresponding author.}%
}

\markboth{Journal of \LaTeX\ Class Files,~Vol.~14, No.~8, August~2021}%
{Shell \MakeLowercase{\textit{et al.}}: A Sample Article Using IEEEtran.cls for IEEE Journals}


\maketitle

\begin{figure*}
    \centering
    \includegraphics[clip,trim={0cm 0.25cm 0cm 0.1cm},width=\linewidth]{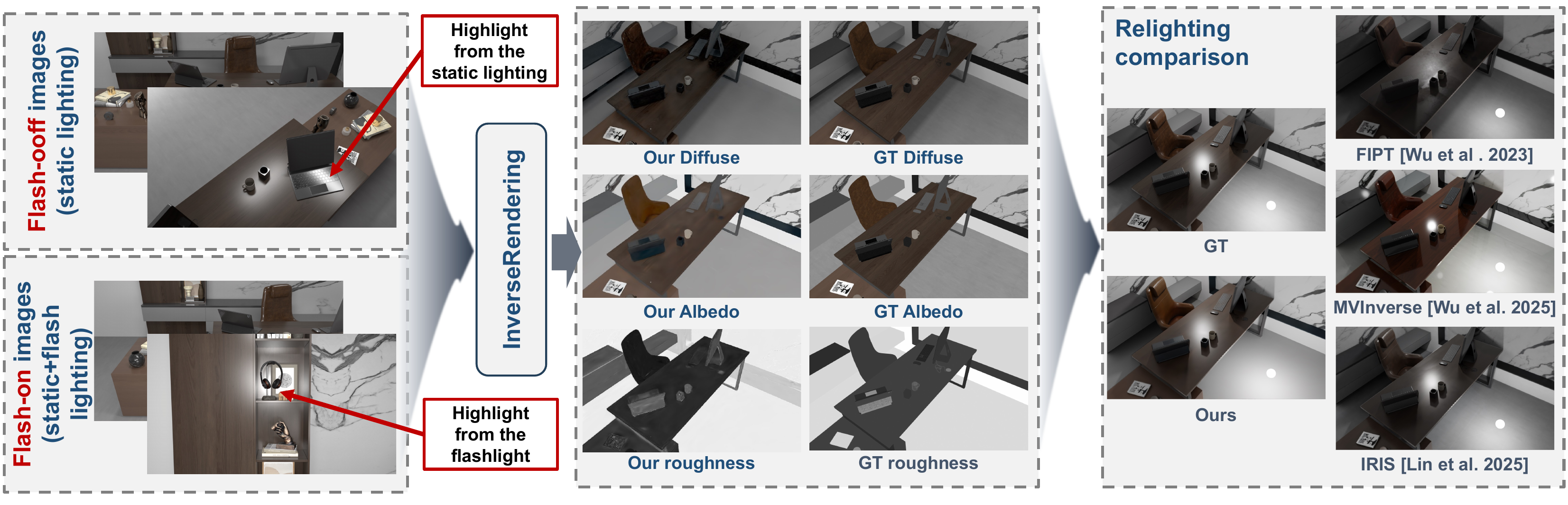}
    \caption{Our method synergizes both static and active lighting information for inverse rendering. By capturing a set of flash-on and flash-off images from different viewpoints, our approach achieves high-quality diffuse–specular decomposition and material estimation, and outperforms current state-of-the-art inverse rendering methods in relighting quality.}
    \label{fig:teaser}
\end{figure*}

\input{sec/0_abstract}    


\input{sec/1_intro}

\input{sec/2_related}
\input{sec/3_method}

\input{sec/4_exp}

\input{sec/5_conclusion}

 

\bibliographystyle{IEEEtran}
\bibliography{main}


 




\vfill

\end{document}

%% file: sec/0_abstract.tex
\begin{abstract}
Recovering geometry, materials, and lighting from photographs is highly ambiguous when only static illumination is available. Active-lighting setups reduce the ambiguity but require dark rooms or specialized hardware. Instead, we synergize both static and flash lighting from casual indoor capture, with the flash on or off, each from independent viewpoints. The flash residual constrains albedo and the BRDF, while static lighting captures grazing-angle specular highlights that flash misses. With a 2DGS reconstruction framing, our key contribution is a GS-anchored diffuse field: a hash-encoded MLP is queried at the rasterized 2DGS depth. As it depends only on world position, it is view consistent in 3D and allows the flash residual to drive material decomposition instead of being absorbed by alpha-blending drift across views. At the same time, we render static lighting with deferred shading such that it can also supervise material decomposition. On five synthetic and three real indoor scenes, our method outperforms six recent baselines on diffuse color, albedo and roughness material parameters, and in relighting where PSNR improves by 4.17 dB over the next-best baseline.
\end{abstract}

%% file: sec/1_intro.tex
\section{Introduction}
\label{sec:intro}

Inverse rendering aims to recover the geometry, materials, and lighting of real-world scenes from image observations. Accurate estimation of these scene properties enables rendering from arbitrary viewpoints under novel lighting and allows editing of physically-based material attributes. However, the task remains highly challenging due to the high degrees of freedom (DoF) and inherent ambiguities in disentangling geometry, materials, and lighting, which complicate the optimization process.

To reduce the degrees of freedom, some works focus on reconstructing objects~\cite{zhang2025ref,yao2025refGS,zhu2025gs}. Here, we can simplify the unknown \emph{static lighting} by modeling it as an environment map or a distant radiance field, and try to estimate it from scratch. However, this is still ill posed and, in scenes with complex lighting, these methods often struggle to disentangle lighting from material. We can also initialize the unknown lighting from captured images and jointly refine the lighting and material~\cite{wu2023factorized,lin2025iris,li2023multi,yu2023milo}, e.g., via explicit path tracing. Such methods can achieve higher-quality material reconstruction if the surrounding captured indirect lighting includes the emitters, but they tend to fail when only part of the scene lighting is observed. A third line of methods use controlled \emph{active lighting} to aid material estimation. Under a one-light-at-a-time (OLAT) setup, they can achieve high-quality materials. To minimize ambient light contamination, they often operate in dark environments using point lights~\cite{iron,kuang2024olat,bi2024gs3,zheng2026ssd}, area lights~\cite{gao2025efficient}, or LED arrays~\cite{ma2021free}, which limits their use in general scenes.

We consider casual capture of indoor environments, and we will use both static and active lighting via flash lighting to achieve higher-quality inverse rendering (Fig.~\ref{fig:teaser}). In our setting, the flash-on and flash-off images can come from different viewpoints and in different numbers. In principle, static and flash lighting sources are \emph{synergistic} in that both provide samples of the material specular lobe. For example, flash lighting can illuminate regions occluded from static lighting, while static lighting captures grazing-angle reflections that are difficult to observe using flash lighting alone. Flash lighting also aids in optimizing albedo, particularly when the static lighting is complex or non-white. Making effective use of these synergistic opportunities requires physically-based reconstruction of both static and active lighting components; e.g., past works isolate direct lighting from flash/no-flash pairs to guide the optimization~\cite{brahimi2024sparse}, or treat static lighting as a free-form residual only rather than a term that must be physically modeled~\cite{cheng2023wildlight}.

Our core contribution to achieve this synergy is a training time spatial structure that lets us constrain geometry and material properties across lighting conditions.
First, consider that we wish to use oriented 2D Gaussian primitives as the underlying representation for reconstruction~\cite{huang20242d} because they are fast to optimize and render.
Such a flash/no-flash optimization is problematic when using a GS method because a) primitives have no natural neighborhood structure, and b) GS representations have many flexible ways to fake view-dependent appearance. For this, we propose a GS-anchored diffuse field, in which the rasterized 2DGS depth at each pixel is used to query a hash-encoded MLP~\cite{muller2022instant} for diffuse color. This design significantly improves diffuse and specular decomposition from optimization because it lets us better leverage static lighting images to assist, and still allows the final result to contain only 2DGS primitives for interactive rendering.
%
In comparison to six current methods, our method achieves state-of-the-art performance in material estimation and relighting, both in metrics and in visual comparisons across eight scenes.

Our contributions are:
\begin{itemize}
\item A method to make synergistic use of casually captured flash and no-flash images of indoor scenes for reconstruction of 2DGS geometry and SVBRDF material parameters. 


\item An optimization method based upon a GS-anchored diffuse field that alleviates the multi-view inconsistency inherent in Gaussian Splatting and improves diffuse and specular decomposition.


\item A set of regularization losses that guide the optimization, resulting in more physically accurate material estimation.
\end{itemize}




%% file: sec/2_related.tex
\section{Related Work}
\label{sec:related}
\paragraph{Inverse rendering under static lighting.}
This aims to recover geometry, spatially varying materials, and unknown lighting from posed images, but remains highly ambiguous due to the entanglement between reflectance and illumination. 
Existing methods can be broadly grouped into object-centric and scene-level approaches. 
For object-centric inverse rendering, recent works have extended 3DGS to relightable reconstruction~\cite{chen2025gi,zhou2025rtr,sun2025svg,choi2024phys3dgs,zhu2025gaussian}. 
For instance, GS-IR~\cite{liang2024gs} combines Gaussian Splatting with deferred physically based shading, using baked visibility and indirect illumination for efficient relighting. 
R3DG~\cite{gao2024relightable} associates radiance and reflectance attributes with Gaussians and performs physically based relighting with ray tracing. 
Follow-up methods further improve reflective appearance modeling and material estimation: Ref-Gaussian~\cite{yao2025refGS} introduces mesh-grounded inter-reflection; Ref-GS~\cite{zhang2025ref} uses directional factorization for view-dependent effects; and MaterialRefGS~\cite{zhangmaterialrefgs} encourages multi-view consistent material inference.
These methods demonstrate the potential of Gaussian representations for static-light inverse rendering, but they commonly rely on simplified or optimized static illumination, making material estimation challenging under complex or partially observed lighting.

For scene-level inverse rendering, the main difficulty lies in modeling complex indoor illumination.
Li et al.~\cite{li2023multi}represent large-scale indoor illumination with HDR textures, while MILO~\cite {yu2023milo} explicitly models multi-bounce light transport in scenes with light-emitting objects.
FIPT~\cite{wu2023factorized} and IRIS~\cite{lin2025iris} further improve scene-level material-lighting estimation with more explicit physically based rendering formulations.
Although these methods can produce high-quality results when scene illumination is sufficiently observed or modeled, they still rely solely on static-light observations, where complex illumination can be easily
absorbed into material appearance.
In contrast, our method complements static-light observations with casually captured flash images, using the active flash residual to provide additional constraints for scene-level Gaussian inverse rendering.


\paragraph{Inverse rendering with active lighting.}
Active-light inverse rendering reduces material-lighting ambiguity by capturing scenes under controlled illumination, such as point lights~\cite{chung2024differentiable}, one-light-at-a-time (OLAT) lighting~\cite{guo2019relightables}, area lights~\cite{zhang2023deep}, or LED arrays~\cite{kang2018efficient}.
IRON~\cite{iron} optimizes neural SDFs and materials from photometric images under point lights, while GS3~\cite{bi2024gs3} and OLAT Gaussians~\cite{kuang2024olat} extend OLAT-style acquisition to Gaussian-based relightable reconstruction.
Other works explore different active-light setups: Gao et al.~\cite{gao2025efficient} use differentiable area-light shading for efficient object reconstruction, and Neural Trace Photography~\cite{ma2021free} uses programmable lighting for free-form scanning of non-planar appearance. 
These methods add constraints for material estimation, but usually require dark-room capture, calibrated lighting devices, or dense active-light measurements, which limit their applicability to casual in-the-wild scenes.

Closest to our setting, WildLight by Cheng et al.~\cite{cheng2023wildlight} constrains the BRDF physically only through the flashlight differential; ambient appearance is modeled by a free-form neural radiance field that takes BRDF parameters as conditioning but is not physically tied to them. As a consequence, strong ambient cues like sunlight from a window producing specular highlights are absorbed by the ambient network rather than entering the BRDF estimate as a physical constraint. Our method instead grounds ambient observations in a physically-based shading model, so both lighting conditions contribute physical evidence to BRDF recovery.


\begin{figure*}[t!]
\centering
\includegraphics[width=1.0\linewidth]{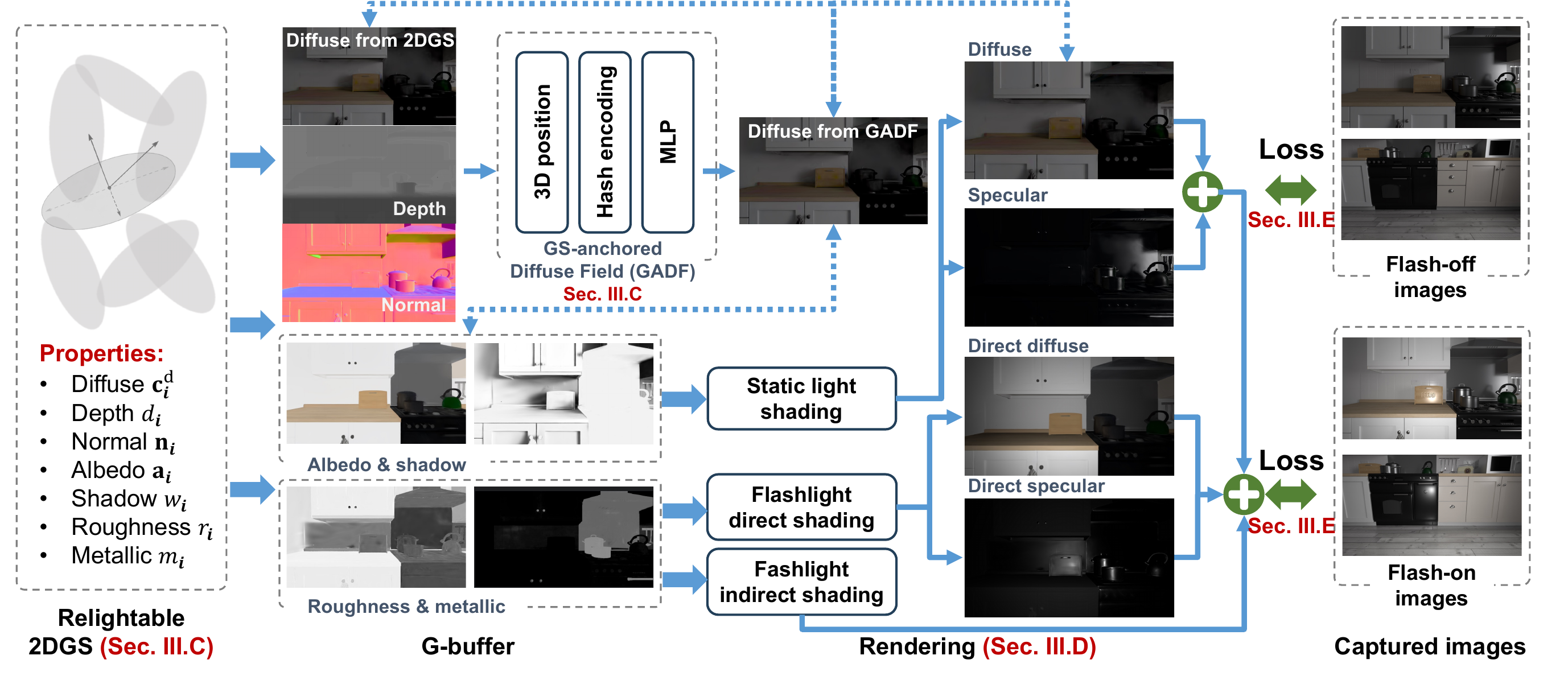}
\caption{\textbf{Our pipeline.} Given a set of relightable 2DGS with material properties, we use deferred rendering to synthesize diffuse and specular colors under static illumination, as well as under a flashlight’s direct and indirect shading components. The rendered results under static and static+flash lighting are then used for inverse rendering by computing a photometric loss. To facilitate diffuse–specular decomposition, we introduce an additional GS-anchored diffuse radiance field (GADF) to regularize the diffuse color in 2DGS. The dashed arrows indicate constraints enforced via loss terms.}
\label{fig:pipeline} 
\end{figure*}

\paragraph{Feed-forward inverse rendering.}
Feed-forward methods predict intrinsic or material properties directly from images, avoiding costly per-scene optimization.
Early intrinsic decomposition methods estimate albedo and shading from a single RGB image~\cite{narihira2015direct}. 
Recent diffusion-based models further improve single-image material decomposition: RGB$\leftrightarrow$X~\cite{zeng2024rgb} predicts intrinsic channels such as albedo, roughness, metallicity, and lighting from indoor images.
This paradigm has also been extended to multi-view settings. 
MVInverse~\cite{wu2025mvinverse} predicts spatially varying materials from multi-view images in a feed-forward manner, while IDArb~\cite{liidarb2025} handles an arbitrary number of input views and illumination conditions with a diffusion model.
DiffusionRenderer~\cite{liang2025diffusion} further leverages video diffusion priors to recover G-buffer attributes from in-the-wild videos. 
These methods are efficient and provide useful material priors, but without scene-specific rendering constraints, their predictions may contain lighting-dependent color casts in albedo and unreliable roughness estimates. 
Therefore, we use feed-forward material predictions as auxiliary priors rather than using them as the final result.

%% file: sec/3_method.tex
\section{Method}
\label{sec:method}

\subsection{Preliminaries}

\paragraph{Gaussian Splatting}

This is a technique for efficient 3D reconstruction and rendering via rasterization~\cite{3dgs}. Each Gaussian is characterized by both geometric attributes $\mu \in \mathbb{R}^3$, covariance matrix $\mathbf{\Sigma} \in \mathbb{R}^{3 \times 3}$, and opacity $o$, as well as appearance attributes $\mathbf{c}$. The Gaussian function for a point $\mathbf{x}$ is defined as:
\begin{align}
\mathcal{G}(\mathbf{x}|\mu, \Sigma) = \exp\left(-\frac{1}{2}(\mathbf{x - \mu})^T\mathbf{\Sigma}^{-1}(\mathbf{x - \mu})\right)
\end{align}
where the covariance matrix $\mathbf{\Sigma}$ is factorized into a rotation matrix $\mathbf{R}$ and a scaling matrix $\mathbf{S}$.

2D Gaussian Splatting (2DGS~\cite{huang20242d}) uses 2D oriented disks as surface elements, in contrast to the ellipsoids used in standard 3DGS, enabling high-quality geometric reconstruction. In 2DGS, Gaussian values are evaluated at the intersections of explicit rays with the 2D disks:
\begin{equation}
\mathcal{G}(\mathbf{u}) = \exp\left(-\frac{u(\mathbf{r})^2+v(\mathbf{r})^2}{2}\right)
\label{eq:gaussian-2d}
\end{equation}
where $\mathbf{u}=(u(\mathbf{r}),v(\mathbf{r}))$ is the intersection point between ray $\mathbf{r}$ and the 2D disk. For rendering, Gaussians are sorted according to their centers and composed into pixels with front-to-back alpha blending:
\begin{equation}
\mathbf{c}(\mathbf{r}) = \sum_{i=1}^N \mathbf{c}_i \alpha_i \prod_{j=1}^{i-1} (1 - \alpha_j), 
\label{eq:alpha_blending}
\end{equation}
where $\alpha_i = o_i \mathcal{G}_i(\mathbf{u})$, which is computed by multiplying the opacity with the Gaussian weight.

\paragraph{Micro-facet BRDF~\cite{cook1982reflectance}}

We follow previous work in inverse rendering for objects and adopt the physically-based rendering (PBR) material model from the Micro-facet BRDF. This allows us to seamlessly import game assets and render our optimized models directly in existing engines without the need for modifications. The model is characterized by three key properties: albedo $\mathbf{a} \in [0,1]^3$, roughness $r \in [0,1]$, and metallicity $m \in [0,1]$. It combines two components: a diffuse term and an isotropic specular GGX lobe~\cite{microfacet}:

\begin{align}
\rho(\mathbf{x}, \omega_{i}, \omega_{o} )&=\frac{(1-m)\mathbf{a}}{\pi}+\frac{D(r)F(m,\mathbf{a})G(r)}{4|\omega_{i} \cdot \mathbf{n}||\omega_{o} \cdot \mathbf{n}|},
\label{eq:brdf}
\end{align}
where $\rho$ is the BRDF, $\mathbf{x}$ represents the shading point, and $\omega_{i}$ and $\omega_{o}$ denote the incident (surface-to-light) and outgoing (surface-to-camera) directions, respectively, and $\mathbf{n}$ is the surface normal. $D$ is the normal distribution function (NDF) that depends on roughness, which uses the GGX model. $F$ is the Fresnel term and $G$ models the shadowing effect between microfacets.

\subsection{Overview}
\label{sec:overview}

We aim to reconstruct both the scene geometry $\mathcal{G}$ and spatially-varying BRDF materials $\mathcal{M}$ from $N_1$ non-flash images $\mathbf{I}^{a}$ and $N_2$ additional flash images $\mathbf{I}^f$. The non-flash images capture static lighting, while the flash images combine both the flash lighting and the static lighting. The flash images are captured at different camera poses than the non-flash images. Our acquisition setup uses a commercial camera with a standard flashlight mounted on top. 

As shown in Fig.~\ref{fig:pipeline}, our inverse rendering pipeline begins by employing a hybrid Gaussian splatting and diffuse field representation (Sec.~\ref{sec:hybrid_rep}) to model the scene’s geometry and materials. Building upon the 2DGS framework, we incorporate material features to each primitive, and use a separate \emph{GS-anchored diffuse field} --- a hash-encoded MLP queried at the rasterized 2DGS depth --- to ensure multi-view consistency of the diffuse color. With this representation, we use deferred rendering for both static and flash lighting scenarios (Sec.~\ref{sec:deferred_shading}) and a multi-stage optimization pipeline with loss terms that guide the inverse rendering process (Sec.~\ref{sec:optimization}).

\paragraph{Why this works.}
The flash provides a known localized perturbation of the lighting at each surface point. Subtracting the rendered non-flash image from the captured flash image isolates this perturbation as a residual $\mathbf{C}_{\rm flash} = \dot{\mathbf{C}} - \mathbf{C}^{\rm nf}$ that separates albedo from incoming radiance. For the residual to drive the decomposition, every other term in the rendering equation must be unable to absorb it. Therefore, the diffuse channel must be a function of world position alone: if the same world point takes different diffuse values from different viewpoints, the optimizer routes the flash signal into that view-dependent capacity instead of the BRDF and lighting terms. While per-Gaussian diffuse parameters are view-consistent in 3D, the effect of rendering many Gaussians from multiple viewpoints is not: a single rasterized pixel mixes contributions from a different set of Gaussians with different alpha-blending weights for each viewpoint, so the diffuse value rasterized at a given world point shifts between views. Under a $1/r^2$ flash, this shift is on the same scale as the residual we want to recover. Thus, routing the diffuse channel through a GS-anchored diffuse field forces difference between the flash and non-flash captures at the same world point to be attributed to the lighting or specular BRDF terms. Forming this as a multi-resolution hash encoding lets the field carry texture-scale detail at fixed parameter cost and with O(1) look-up.

\subsection{GS-anchored Diffuse Field for Diffuse-Specular Decomposition}
\label{sec:hybrid_rep}

\begin{figure}[t!]
\centering
\includegraphics[width=1.0\linewidth]{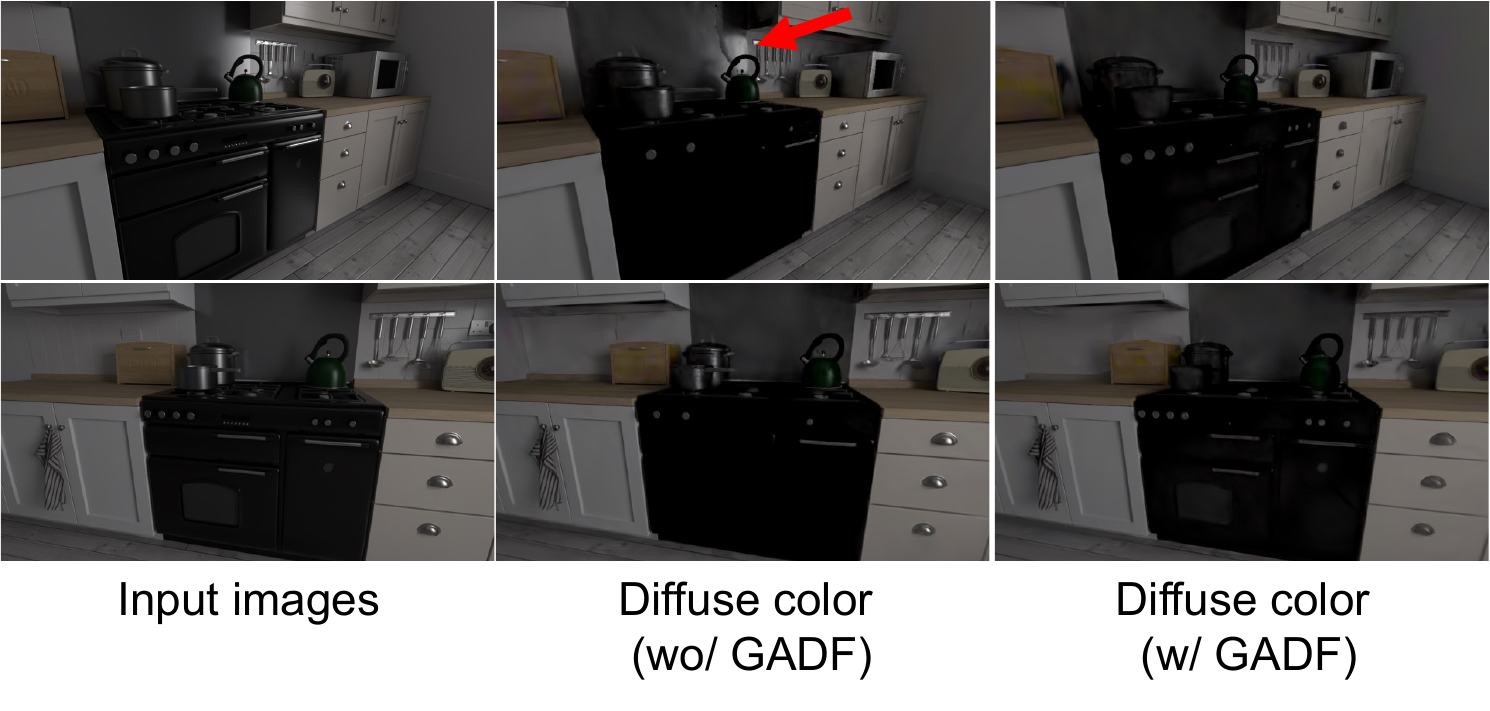}
\caption{Our GS-anchored diffuse field constrains the diffuse color in 2DGS, reducing view-dependent effects such as highlights (red arrow) and improving the decomposition of diffuse and specular components.}
\label{fig:GADF} 
\end{figure}

Our scene representation extends the 2D Gaussian Splatting (2DGS) method by incorporating additional material parameters. For each 2D Gaussian, we include not only its position, rotation, and opacity, but also the diffuse color $\mathbf{c}^\mathrm{d}$, BRDF parameters $\mathbf{a}$, $r$, $m$, and shadow $w$. We treat the diffuse color $\mathbf{c}^\mathrm{d}$ as a learnable parameter, which guides the decomposition of albedo and shadow from the lighting. These per-primitive attributes are alpha-blended into image-space buffers $\mathbf{C}^\mathrm{d}, \mathbf{A}, R, M, W, \mathbf{N}, D$ for the shading pass (Sec.~\ref{sec:deferred_shading}).

To disentangle the diffuse component from the color, for diffuse color $\mathbf{c}^\mathrm{d} \in [0,1]^3$, we remove the spherical harmonics (SH). A similar approach is used in Ref-Gaussian~\cite{yao2025refGS} and MaterialRefGS~\cite{zhangmaterialrefgs}, which assume that specular variations can be decomposed into view-independent material properties, while the final view-dependent appearance is governed by the BRDF. However, since Gaussians span different pixels and contribute varying weights across viewpoints during alpha blending, material parameters exhibit significant inconsistencies from different perspectives. This inconsistency complicates accurate illumination decomposition, as the BRDF struggles to infer a consistent global physical reflectance effect from such varying material observations.

To address this issue, we propose a GS-anchored diffuse field that takes the 3D position from the 2DGS rendering as input to enforce multi-view consistency for the diffuse color. For each view, we first retrieve the depth from rasterization and then compute the estimated diffuse color $\hat{\mathbf{C}}^\mathrm{d}$:
\begin{align}
\hat{\mathbf{C}}^\mathrm{d} = \mathrm{Sigmoid}(\mathrm{MLP_{df}}(\mathbf{x}({D)}))
\end{align}
where $\mathrm{MLP_{df}}$ uses hash encoding. Here, $D(\mathbf{u})$ denotes the alpha-blended depth rendered by 2DGS using the same front-to-back compositing weights as in Eq.~\eqref{eq:alpha_blending}. We unproject pixel $\mathbf{u}$ with $D(\mathbf{u})$ to obtain the world-space position queried by the GADF. Unlike MaterialRefGS, which uses a warping loss between neighboring views to improve multi-view consistency, our GS-anchored diffuse field implicitly ensures view consistency. As shown in Fig.~\ref{fig:GADF}, this design leads to more accurate diffuse and specular decomposition. This solution relies on the rendered depth being multi-view consistent; we achieve this through a depth distortion loss that pushes primitives together (e.g., towards true surfaces), and a normal prior loss. This implies that any inconsistency of diffuse color arises not from the virtual geometry but from the alpha blending process. The architecture and hash-grid parameters are provided in the supplementary material.

Although the other G-buffer attributes are also view-independent in principle, we use the GADF only to represent diffuse color. In our experiments, predicting all view-independent material properties with the GADF resulted in a loss of fine-scale details. We attribute this degradation to small inconsistencies in the world-space positions queried from the rasterized depth across different views. These inconsistencies become more detrimental
when multiple correlated material channels are predicted from the same spatial field. We therefore retain per-Gaussian representations for the remaining material properties and use the GADF only for diffuse color. We validate this design choice in the ablation study.

Additionally, the diffuse color $\mathbf{C}^\mathrm{d}$ contains contributions from both lighting and shadow. Unlike Ref-Gaussian and MaterialRefGS, which treat diffuse color as the ``base color'', we further disentangle it into albedo, lighting, and shadow via the constraint introduced in Sec.~\ref{sec:deferred_shading} and the loss in Sec.~\ref{sec:optimization}.

\subsection{Deferred Shading under Static and Flash Lighting}
\label{sec:deferred_shading}

\paragraph{Static Lighting} 

We model the static incident radiance using a set of spatially varying environment maps. Specifically, we partition the scene bounding box into a regular grid of cubic cells, whose size is determined by the estimated scene height, and associate an environment map with the center of each cell. For a shading point \(\mathbf{x}\), we query the environment maps at the eight neighboring grid cells and obtain the local incident illumination by trilinearly interpolating their lighting contributions according to the position of \(\mathbf{x}\) within the grid.


The shading under static lighting $\mathbf{C}^\mathrm{nf}$ consists of the diffuse and specular components:
\begin{align}
    \mathbf{C}^\mathrm{nf}(\omega_o) &= \mathbf{C}^\mathrm{d}(\omega_o) + \mathbf{C}^\mathrm{s}(\omega_o).
\label{equ:c_nf}
\end{align}

For the diffuse component $\mathbf{C}^\mathrm{d}$, as described in the previous section, it is derived by rasterizing the diffuse color of each 2D Gaussian into the image space. We factor the diffuse color into the albedo and the diffuse environment map, along with the shadow, by applying a constraint using a L$1$ loss, denoted as $\leftrightarrow$:
\begin{align}
    \mathbf{C}^\mathrm{d} &\leftrightarrow (1-M)\mathbf{A} \int_\Omega L(\omega_i)\frac{(\omega_i \cdot \mathbf{N})}{\pi} d \omega_i \\
    &\approx (1-M)\mathbf{A} \cdot L^\mathrm{env}_\mathrm{diffuse}(\mathbf{N}) \cdot W,
\end{align}
where $L_{\mathrm{diffuse}}^{\mathrm{env}}$ denotes the diffuse irradiance map obtained by convolving the raw environment map with the cosine kernel, and $\mathbf{W}$ is the image-space shadow map.

To model shadowing, we associate each Gaussian with a learnable scalar shadow attribute $w_i$, which is alpha-blended during rasterization to produce the image-space shadow map $\mathbf{W}$. The shadow attributes are jointly optimized through the photometric rendering loss and the diffuse decomposition loss in Eq.~\eqref{equ:loss_d}. We deliberately avoid directly supervising $\mathbf{W}$ with shadow rays traced against the reconstructed 2DGS geometry, since visibility estimates derived from incomplete or approximate Gaussian geometry can be unreliable and may introduce erroneous supervision. Instead, the flash-induced residual provides the primary cue for disentangling albedo from shadowing.

For the specular component $\mathbf{C}^\mathrm{s}$, we follow recent works using the split-sum approximation~\cite{munkberg2022extracting} to factorize the integral into a specular lobe term $L_\mathrm{specular}$, and a BRDF integral term $F_\mathrm{specular}$:
\begin{align}
     \mathbf{C}^\mathrm{s}(\omega_o) &= \int_\Omega L(\omega_i) \frac{D F G}{4 (\omega_o \cdot \mathbf{N})} d \omega_i \\
     &\approx L_\mathrm{specular} \cdot F_\mathrm{specular}
    \label{eq:specular}
\end{align}
where $L_\mathrm{specular} = L_\mathrm{specular}^{env}(\omega_r,R)$ is a learnable environment map with Mip-map, which is queried using the reflected ray $\omega_r$ and roughness. $F_\mathrm{specular}$ can be computed directly as $F_\mathrm{specular} = ((1 - M)0.04 + M * \mathbf{A}) * F_1 + F_2$, where $F_1$ and $F_2$ are pre-computed scalars. 

We employ separate environment-map representations for the diffuse and specular illumination. The two representations are optimized independently rather than being derived from a shared underlying environment map. Specifically, the diffuse environment map is queried using the surface normal $\mathbf{N}$, whereas the mipmapped specular environment map is queried using the reflected direction $\omega_{r}$ and surface roughness $R$, allowing the two components to capture their respective low- and high-frequency illumination characteristics.

\paragraph{Flash lighting} 

In the flash case, the captured radiance $\mathbf{C}^\mathrm{f}$ consists of the static lighting term $\mathbf{C}^\mathrm{nf}$ plus an additional flashlight term induced by the auxiliary flash:
\begin{align}
\mathbf{C}^\mathrm{f}(\omega_{o}) = \mathbf{C}^\mathrm{nf}(\omega_{o}) + \left(\mathbf{C}_{\rm{dir}}^\mathrm{af}(\omega_{o})+\mathbf{C}_{\rm{ind}}^\mathrm{af}(\omega_{o})\right).
\label{eq:color}
\end{align}
where $\mathbf{C}_{\rm dir}^\mathrm{af}$ and $\mathbf{C}_{\rm ind}^\mathrm{af}$ denote the direct and indirect shading contributions of the flashlight, respectively.

To reduce the complexity of optimization, we approximate the flashlight as a point light source colocated with the camera, ignoring its positional offset relative to the camera for both direct and indirect shading. Under this approximation, we treat the two shading components differently. For direct shading, the flash illumination is evaluated explicitly using the colocated point-light model, which simplifies the lighting formulation during optimization. For indirect shading, to further improve computational efficiency, we follow FIPT~\cite{wu2023factorized} and factorize the diffuse and specular contributions of the flash illumination. The resulting shading terms are pre-baked and subsequently queried and interpolated during optimization.

Unlike a conventional point light, we model the direct radiance distribution of the flashlight using a multi-layer perceptron (MLP), which attenuates the intensity isotropically in directions perpendicular to the centerline. In the global coordinate system, its position is denoted by $\mathbf{x}^l$, and its centerline direction is defined as $\omega_l$. The direct lighting color from the flashlight observed at a point $\mathbf{x}$ in the direction $\omega_o$ can then be computed as:
\begin{align}
\mathbf{C}_{\rm{dir}}^\mathrm{af} &= \frac{\operatorname{MLP}_{\theta}
\left(
\boldsymbol{\omega}_{\mathbf{x}_l \rightarrow \mathbf{x}} \cdot
\boldsymbol{\omega}_l
\right)}{\|\mathbf{x}-\mathbf{x}^l\|_2^2} \\ &\cdot
\rho\left(\mathbf{x},
\boldsymbol{\omega}_{\mathbf{x}\rightarrow\mathbf{x}_l},
\boldsymbol{\omega}_o
\right)
\cdot 
\max\left(
\mathbf{n} \cdot
\boldsymbol{\omega}_{\mathbf{x}\rightarrow\mathbf{x}_l},
0
\right),
\label{eq:flash_color}
\end{align}
where $\rho(\mathbf{x},\boldsymbol{\omega}_i,\boldsymbol{\omega}_o)$
is the complete microfacet BRDF defined in Eq.~\eqref{eq:brdf},
including both its diffuse and specular components.


The indirect lighting is computed as the integral of the radiance induced by the flash:
\begin{align}
    \mathbf{C}_\mathrm{ind}^\mathrm{af}
    &=\int_{\Omega^+} L_\mathrm{ind}(\mathbf{x},{\omega}_i)
    \rho (\mathbf{x},\omega_{i}, \omega_{o} )\max\left(
\mathbf{n} \cdot {\omega}_{i}, 0
\right),d{\omega}_i,
\end{align}
where $\rho$ is the BRDF.

To improve computational efficiency, we follow the factorized precomputation strategy of FIPT~\cite{wu2023factorized}, which separates the BRDF-dependent terms from the incident illumination and pre-bakes the resulting shading components using Monte Carlo integration for efficient reuse during optimization. In the original FIPT formulation, the indirect illumination is decomposed into one diffuse shading term, $L_\mathrm{d}$, and two specular shading terms, $L_\mathrm{s}^0$ and $L_\mathrm{s}^1$, corresponding to the two Fresnel-related terms under the Schlick approximation~\cite{schlick1994inexpensive}.

Our setting is considerably simpler than that of FIPT, as we model only the additional illumination induced by the camera flash rather than general indoor illumination. We therefore simplify the factorization to further reduce the precomputation cost. In particular, we discard indirect radiance originating from regions outside the field of view of the corresponding flash image and retain only the diffuse component of the indirect flash illumination. Accordingly, we assume that the indirect radiance induced by the flash is purely diffuse and omit the specular indirect terms.

For a shading point $\mathbf{x}$, we estimate the incident indirect flash radiance along a sampled direction $\omega_i^{(k)}$ as
\begin{align}
L_\mathrm{ind}(\mathbf{x},\omega_i^{(k)}) &\approx \mathbf{C}_{\rm{flash}}(\mathrm{proj}(\mathcal{M}(\mathbf{x},\omega_i^{(k)})),\\
\mathbf{C}_{\rm{flash}} &= \dot{\mathbf{C}} - \mathbf{C}^{\mathrm{nf}},
\end{align}
where $\omega_i^{(k)}$ denotes the direction of the $k$-th sampled ray originating from $\mathbf{x}$. The ray is intersected with the reconstructed mesh $\mathcal{M}$, and the resulting surface point is projected onto the corresponding flash image to retrieve the flash-induced radiance. Here, $\dot{\mathbf{C}}$ denotes the captured image radiance under both static and flash illumination, whereas $\mathbf{C}^{\mathrm{nf}}$ is the rendered non-flash radiance. Their difference therefore provides an estimate of the radiance contribution induced by the flash.

The mesh $\mathcal{M}$ is extracted from the optimized 2DGS geometry after the static-light optimization stage and remains fixed throughout the subsequent flash-light optimization. For each sampled ray originating from $\mathbf{x}$ along $\omega_i^{(k)}$, we first compute its intersection with $\mathcal{M}$. The operator $\operatorname{proj}(\cdot)$ then projects the intersection point into the corresponding flash image using the calibrated camera intrinsics and extrinsics, where the flash residual $\mathbf{C}_{\mathrm{flash}}$ is retrieved through bilinear interpolation. Samples whose projected locations fall outside the image domain are ignored.

\subsection{Optimization}
\label{sec:optimization}

We use two stages. First, we optimize the scene with only non-flash images, and then we add the flash images with a batch size of two: one non-flash and one flash image.

We define a loss $\mathcal{L}$ as the weighted combination of the image loss and a set of regularizer terms:
\begin{equation}
    \mathcal{L} = \mathcal{L}_{\rm r} + \mathcal{L}_{\rm p} +
    \mathcal{L}_{\rm d} + 
    \lambda_{\rm s} \mathcal{L}_{\rm s} + 
    \lambda_{\rm sem} \mathcal{L}_{\rm sem} + 
    \lambda_{\rm 2dgs} \mathcal{L}_{\rm 2dgs},
    \label{eq:loss}
\end{equation}
where $\lambda$ is the weight of each term.

\paragraph{Rendering loss}
This measures the difference between captured pixel colors $\dot{\mathbf{C}}$ and rendered pixel colors $\mathbf{C}^\mathrm{nf}$ or $\mathbf{C}^\mathrm{f}$:
\begin{align}
\mathcal{L}_\mathrm{r} = (1-m)\lVert\dot{\mathbf{C}} - \mathbf{C}^\mathrm{nf}\rVert_1+m\lVert\dot{\mathbf{C}} - \mathbf{C}^\mathrm{f}\rVert_1.
\end{align}
where $m \in \{0, 1\}$ indicates when the image contains flash lighting.

\paragraph{Prior loss}
Similar to other methods, we use surface normals as priors to regularize the 2DGS reconstruction. These normals are computed using MVInverse~\cite{wu2025mvinverse}, which performs multiview material and normal estimation. Since the normals from MVInverse are often inaccurate near geometric boundaries, we extract an edge map from the normal map of each image to identify regions of normal discontinuity. This edge map is then used to generate a weight map, $\mathbf{W}_\mathrm{eg}$, which reduces the influence of the regularization near edges.  

Furthermore, to improve the ill-posed albedo and shadow optimization, we introduce an albedo prior loss, where $\nabla$ denotes the gradient along the horizontal and vertical directions:
\begin{align}
\mathcal{L}_\mathrm{p} = \lambda_\mathrm{n}\mathbf{W}_\mathrm{eg}\lVert \mathbf{N} - \tilde{\mathbf{N}}\rVert_1+
\lambda_\mathrm{a}\lVert \mathbf{A} - \tilde{\mathbf{A}} \rVert_1+\lambda_\mathrm{g}\lVert \nabla\mathbf{A} - \nabla\tilde{\mathbf{A}} \rVert_1.
\end{align}
Here, $\tilde{\mathbf{A}}$ and $\tilde{\mathbf{N}}$ denote the albedo and normal maps estimated by MVInverse. We do not use the roughness and metallic maps from MVInverse, as their quality is insufficient for our purposes. We only apply the prior loss to the non-flash image in each batch, as MVInverse is primarily designed for images captured under static lighting.

\paragraph{Diffuse decomposition loss}
We penalize the rendered diffuse color $\mathbf{C}^\mathrm{d}$ against the diffuse color from the GS-anchored diffuse field $\hat{\mathbf{C}}^\mathrm{d}$. The rendered diffuse color is further decomposed into the metallic, albedo, diffuse irradiance, and shadow maps, using an $L_1$ loss:
\begin{align}
\mathcal{L}_\mathrm{d} &= \lambda_{\rm{dd}}\lVert \mathbf{C}^\mathrm{d} - \hat{\mathbf{C}}^\mathrm{d}\rVert_1 \\
&+ 
\lambda_{\rm{da}} \lVert
\mathbf{C}^\mathrm{d} - (1-M)\mathbf{A} \cdot L^\mathrm{env}_\mathrm{diffuse}(\mathbf{N}) \cdot W
\rVert_1.
\label{equ:loss_d}
\end{align}

\paragraph{Specular decomposition loss}
To improve the estimation of the roughness parameter from specular reflections, we compute the residual between the captured image and the sum of the non-flash color and the diffuse shading component of the flash illumination $\mathbf{R}_\mathrm{sp}$. We then use this residual to identify low-roughness regions and apply a regularization constraint to these segmented regions, since strong specular highlights typically occur in such areas:
\begin{align}
\mathcal{L}_\mathrm{s} &= \mathrm{SEM}(\mathbf{R}_\mathrm{sp} > 0.5) \lVert R-0.05\rVert_1,\\
\mathbf{R}_\mathrm{sp}&=\dot{\mathbf{C}}-\left(\mathbf{C}^\mathrm{nf}+\mathcal{D}_{\mathrm{diff}}(\mathbf{C}^\mathrm{af}_\mathrm{dir}) + \mathcal{D}_{\mathrm{diff}}(\mathbf{C}_\mathrm{ind}^\mathrm{af})\right),
\end{align}
where $\mathrm{SEM}(\cdot)$ denotes the semantic segmentation mask for a specific region, represented as a binary mask. $\mathcal{D}_{\mathrm{diff}}(\cdot)$ denotes the operator
that extracts the diffuse component from the flash-shading contribution.

We also include a semantic loss, $\mathcal{L}_{\mathrm{sem}}$, as used in FIPT, along with the regularization losses from 2DGS, denoted by $\mathcal{L}_{\mathrm{2dgs}}$. Please refer to the supplementary material for additional details.

\begin{table*}[t]
\centering
\caption{Quantitative comparison. Blank entries indicate unavailable results.}
\label{tab:comparison_without_scene}
\small
\setlength{\tabcolsep}{3pt}
\begin{tabular}{lcccccccccccc}
\toprule
\multirow{2}{*}{Method} 
& \multicolumn{3}{c}{Diffuse Color} 
& \multicolumn{3}{c}{Relighting} 
& \multicolumn{3}{c}{Albedo} 
& \multicolumn{1}{c}{Roughness}
& \multicolumn{1}{c}{Normal}
& \multirow{2}{*}{Runtime} \\
\cmidrule(lr){2-4}
\cmidrule(lr){5-7}
\cmidrule(lr){8-10}
\cmidrule(lr){11-11}
\cmidrule(lr){12-12}
& PSNR $\uparrow$ & SSIM $\uparrow$ & LPIPS $\downarrow$
& PSNR $\uparrow$ & SSIM $\uparrow$ & LPIPS $\downarrow$
& PSNR $\uparrow$ & SSIM $\uparrow$ & LPIPS $\downarrow$
& MAE $\downarrow$ & MAE $\downarrow$ & \\
\midrule

Ours          
& \textbf{27.49} & \textbf{0.852} & \textbf{0.151} 
& \textbf{25.89} & \textbf{0.916} & \textbf{0.147} 
& \textbf{24.14} & \textbf{0.856} & 0.178 
& \textbf{0.1554} & \textbf{0.0211} & 40 min \\

Ours-SAM2      
& 26.85 & 0.838 & 0.185 
& 24.38 & 0.895 & 0.166
& 22.96 & 0.843 & \textbf{0.174} 
& 0.1942 & 0.0258 & 40 min \\

FIPT~\cite{wu2023factorized}          
& -- & -- & -- 
& 20.96 & 0.888 & 0.180 
& 19.65 & 0.780 & 0.207 
& 0.2794 & -- & 95 min \\

IRIS~\cite{lin2025iris}          
& -- & -- & -- 
& 21.72 & 0.877 & 0.173 
& 20.63 & 0.799 & 0.221 
& 0.1848 & -- & 95 min \\

WildLight~\cite{cheng2023wildlight}
& -- & -- & --
& 20.11 & 0.794 & 0.331
& 16.96 & 0.640 & 0.507
& 0.2536 & 0.0758 & 480 min \\

MaterialRefGS~\cite{zhangmaterialrefgs} 
& 17.28 & 0.665 & 0.288 
& -- & -- & -- 
& -- & -- & -- 
& 0.2760 & 0.0287 & 110 min \\

Ref-Gaussian~\cite{yao2025refGS}  
& 24.25 & 0.787 & 0.175 
& -- & -- & -- 
& -- & -- & -- 
& 0.3006 & 0.0332 & 25 min \\

MVInverse~\cite{wu2025mvinverse}     
& -- & -- & -- 
& 21.69 & 0.883 & 0.209 
& 20.45 & 0.848 & 0.211 
& 0.2934 & 0.0341 & 3 min \\

\bottomrule
\end{tabular}
\end{table*}

\section{Implementation Details}  
\label{sec:imp}


For all experiments, each scene is trained for 25,000 iterations on a single NVIDIA RTX 5090 GPU. Training is carried out in two stages. In the first stage, we optimize the scene geometry, material properties, and multiple environment maps for 18,000 iterations, utilizing static illumination only. In the second stage, covering iterations 18,000 to 25,000, we enable the flashlight for material optimization, simultaneously refining the material properties, environmental lighting, and flashlight parameters. During the second stage, we set the batch size to two: for each batch, we first sample a non-flash image and then select the nearest flash image based on the camera’s position and orientation. All methods are trained and tested using images with a resolution of $640 \times 360$.

During optimization, we use the Adam optimizer~\cite{kingma2014adam} for both the 2DGS and material parameters, following the same learning rates as Ref-Gaussian~\cite{yao2025refGS}. For the GS-anchored diffuse field, implemented using an Instant-NGP~\cite{muller2022instant} network, we use Adam with an initial learning rate of $0.001$ and a delay multiplier of $0.1$, which gradually reduces the learning rate to $0.0001$. The loss-term weights are set to $\lambda_{\rm n}=0.5$, $\lambda_{\rm a}=0.05$, $\lambda_{\rm g}=2.0$, $\lambda_{\rm {dd}}=2.0$, $\lambda_{\rm {da}}=0.05$, and $\lambda_{\rm s}=0.3$. For the albedo-gradient prior, we use $\lambda_{\rm g}=2.0$ for all synthetic scenes and $\lambda_{\rm g}=10.0$ for all real-world scenes. The larger weight for real-world captures is used to compensate for the higher level of capture noise and reconstruction uncertainty. We use the same $\lambda_{\rm g}$ for all scenes within each category and do not tune it separately for individual scenes.

\paragraph{Real image capture}
For real-scene experiments, we use a Sony A7M4 camera equipped with a Godox V350S flash. For each scene, we keep the camera exposure settings, white balance, and flash power fixed across all captures, and disable automatic exposure and automatic white-balance adjustments to ensure a consistent photometric response between the flash and non-flash images. The white balance is set to 5500K. All captured images are stored as 8-bit sRGB PNG images. Our shading model computes rendered colors in linear RGB space. Before computing the photometric loss, we convert the rendered linear RGB colors to sRGB so that they are compared with the captured images in the same color space. We do not explicitly estimate a camera response function or mask saturated pixels during optimization.

During optimization, we assume that the flash is collocated with the camera. Accordingly, for each flash image, we set the flash position $\mathbf{x}_l$ to the corresponding camera center and align the flash centerline direction $\boldsymbol{\omega}_l$ with the camera optical axis. We do not explicitly calibrate the camera-to-flash extrinsic transformation. The robustness to violations of this collocation assumption is evaluated in the ablation study.

%% file: sec/4_exp.tex
\section{Experiments}
\label{sec:experiments}


\paragraph{Datasets}

We evaluate our method on both synthetic and real-world indoor scenes with unpaired flash and non-flash observations. Our synthetic dataset contains five Blender-rendered scenes, including Kitchen, Office, Market, Livingroom-1, and Livingroom-2. For each synthetic scene, we render 100--150 non-flash images and 100--150 flash images at a resolution of $1280 \times 720$. Since the synthetic scenes are generated in Blender, ground-truth material maps are available and used for quantitative material evaluation.

We further capture three real-world scenes, including Coffee bar, Classroom, and Car, using a Sony A7M4 camera equipped with a Godox V350S flash. Each real scene contains 80--100 non-flash images and 80--100 flash images. The camera poses of real scenes are estimated using COLMAP. For all scenes, we use MVInverse~\cite{wu2025mvinverse} to generate per-view albedo maps and normal maps as auxiliary supervision during training. For real-world scenes, we mainly evaluate material quality through qualitative decomposition and relighting results.

\paragraph{Evaluation Metrics}

We evaluate image reconstruction quality using PSNR, SSIM, and LPIPS. For synthetic scenes, we assess material estimation using the available ground-truth material maps. For relighting evaluation, we select five sample images and illuminate each with five different point lights as well as an environment map. We then compute the metrics by comparing the rendered relighting images—generated using our G-buffer and SSRT—with the corresponding ground-truth relighting images. For real-world scenes, where ground-truth materials are unavailable, we focus on qualitative evaluation, assessing material plausibility through decomposition results and relighting performance.

\begin{figure*}[t!]
\centering
\includegraphics[width=0.95\linewidth]{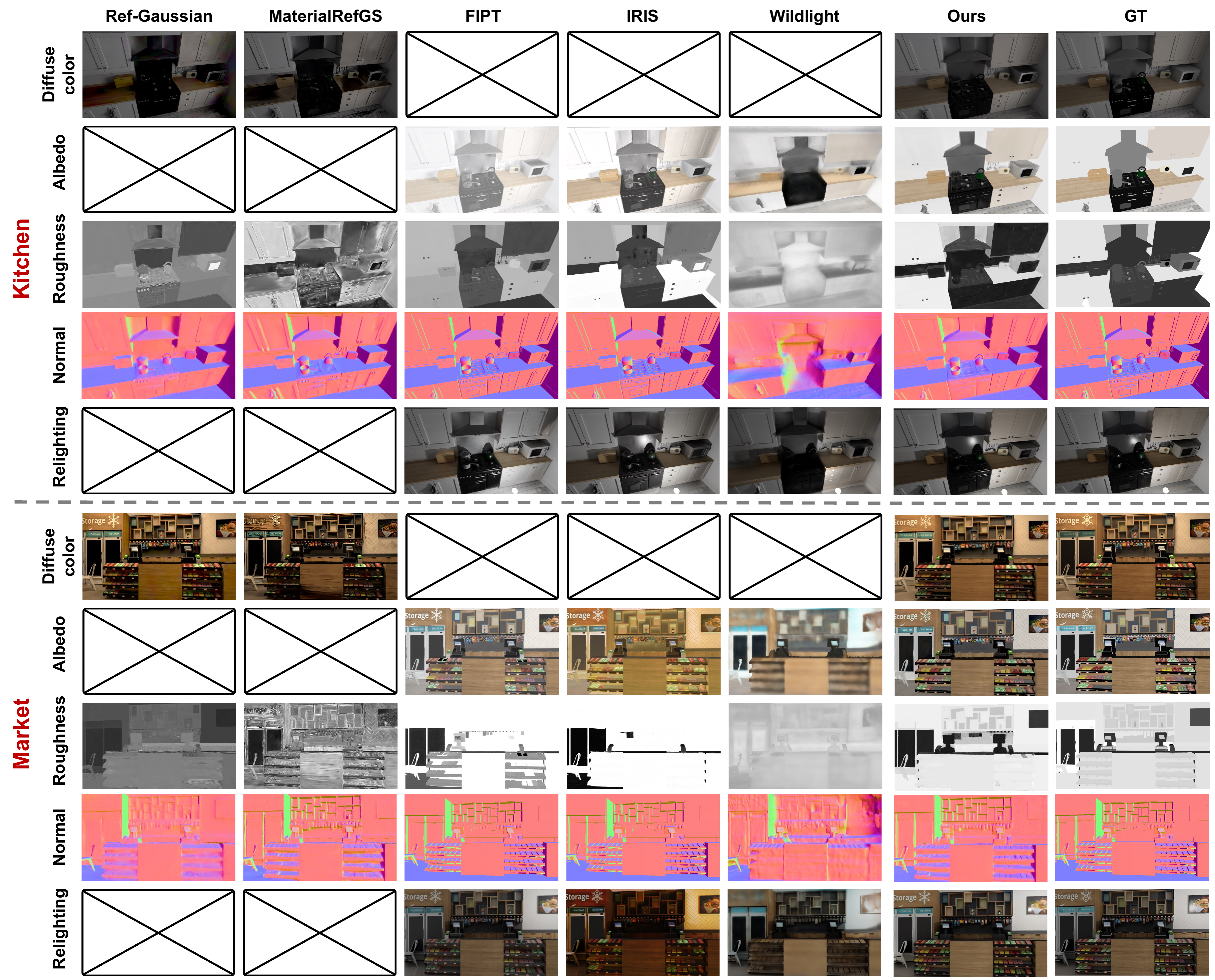}
\caption{\textbf{Comparison of synthetic scenes (1/2).} 
Qualitative comparison of material decomposition and relighting. 
For the FIPT and IRIS experiments, which utilize the ground-truth mesh, 
we only present the ground-truth normals.}
\label{fig:comparison-1-1} 
\end{figure*}

\begin{figure*}[t!]
\centering
\includegraphics[width=0.95\linewidth]{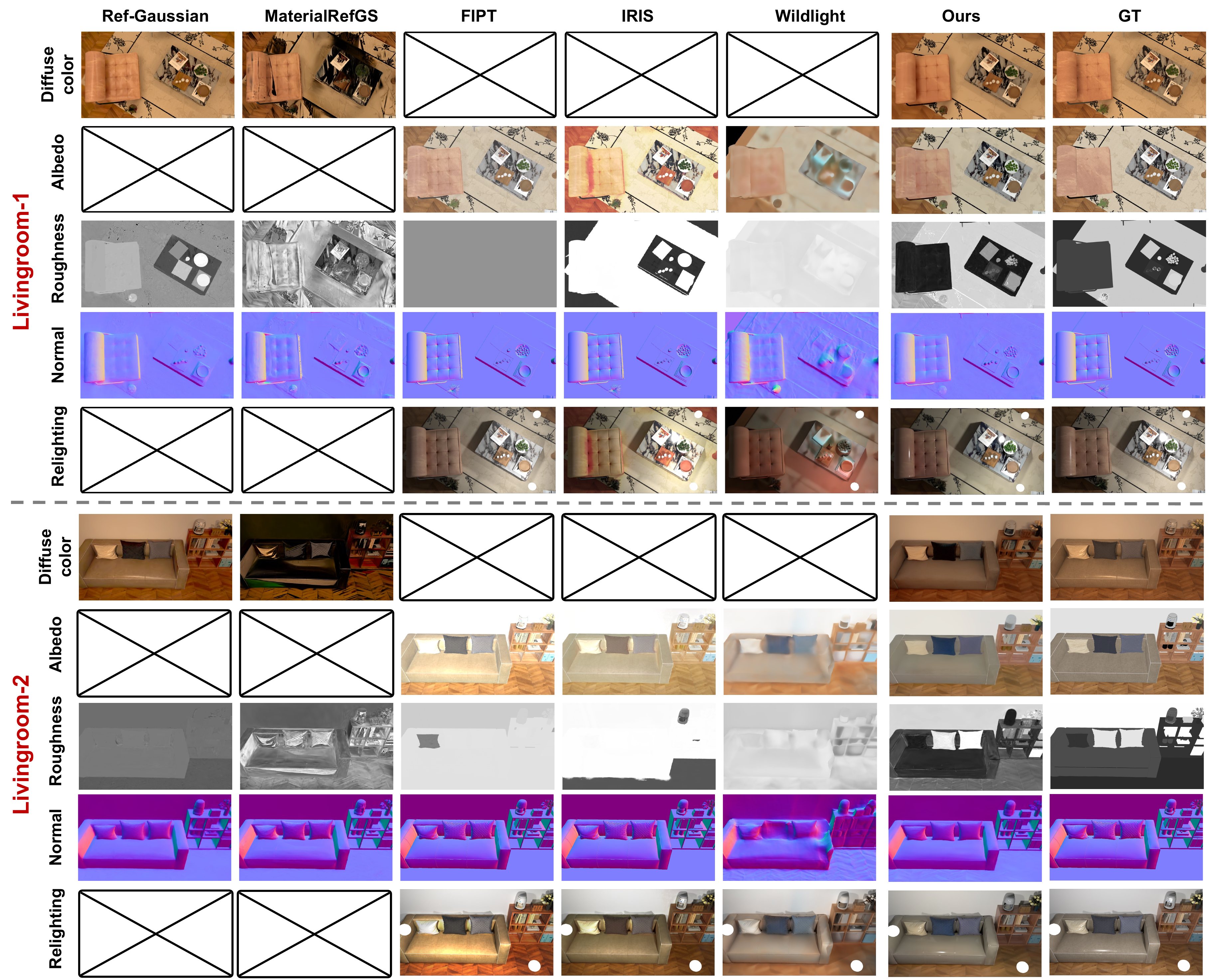}
\caption{\textbf{Comparison of synthetic scenes (2/2).} 
Continued from Fig.~\ref{fig:comparison-1-1}.}
\label{fig:comparison-1-2} 
\end{figure*}

\paragraph{Baselines}

\begin{figure*}[t!]
\centering
\includegraphics[width=0.9\linewidth]{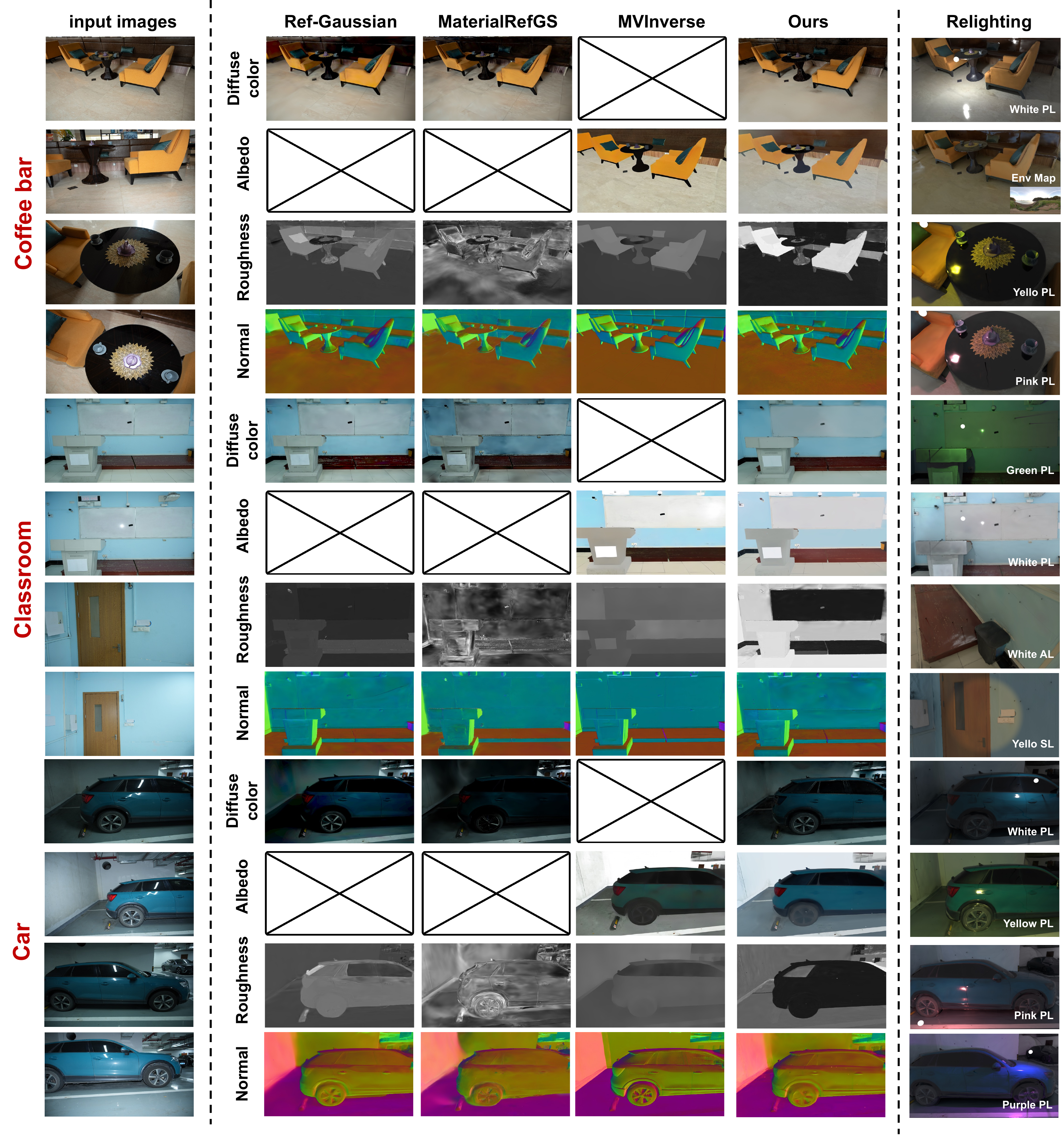}
\caption{\textbf{Comparison of real-world scenes.} For the relighting results, we report both the color and type of lighting: PL for point light, AL for area light, and SL for spotlight. Note that the original light ball may appear white, but its radiance is not truly white due to overexposure when observing the light source directly. Please refer to the video for additional visual comparisons.}
\label{fig:comparison-2} 
\end{figure*}

We compare our method with several representative inverse rendering approaches under different reconstruction paradigms. Since WildLight~\cite{cheng2023wildlight} also exploits active illumination for inverse rendering, it is particularly relevant to our acquisition setting. We therefore include WildLight as a baseline and compare its material decomposition and relighting results with those of our method under similar input conditions.

We further compare with Gaussian-based reflective reconstruction methods, including Ref-Gaussian~\cite{yao2025refGS} and MaterialRefGS~\cite{zhangmaterialrefgs}. For a fair comparison, both methods are provided with ground-truth segmentation masks and the corresponding segmentation loss. Since MaterialRefGS already incorporates its own normal prior loss, we additionally apply the same normal prior loss used in our method to Ref-Gaussian.

We also evaluate two path-tracing-based inverse rendering methods, FIPT~\cite{wu2023factorized} and IRIS~\cite{lin2025iris}, both of which require explicit geometry. We therefore provide them with the ground-truth mesh and segmentation masks. FIPT additionally requires HDR input. To satisfy the input requirements of these methods and ensure sufficient coverage of the scene and light source, we supplement our captured images with approximately 50 additional views.

Finally, MVInverse is included as a feed-forward multi-view inverse rendering baseline. To demonstrate that our method does not rely on ground-truth 2D segmentation, we additionally conduct experiments using per-view segmentation masks generated by SAM2~\cite{ravi2025sam}.

\subsection{Comparisons}
\label{sec:comparisons}

\paragraph{Synthetic scenes}

We first evaluate our method on five synthetic scenes with available ground-truth material maps. Since the compared methods support different output representations and target different tasks, we evaluate each method based on the predictions it produces. For Ref-Gaussian and MaterialRefGS, we compare the estimated diffuse color, roughness, and normal maps. For WildLight, MVInverse, FIPT, and IRIS, we compare the estimated albedo and roughness maps, as well as the relighting results.

For relighting, we use Blender's SSRT renderer with the per-view G-buffers predicted by our method. We adopt SSRT for its efficiency and compatibility with Blender. Since our method does not reconstruct complete geometry, ray tracing may introduce artifacts in regions with incomplete geometry. In contrast, SSRT operates directly on the rendered G-buffers and is therefore more robust to such incompleteness.

Table~\ref{tab:comparison_without_scene} presents the quantitative comparison of our method with the baselines on the synthetic dataset. Our method achieves more accurate diffuse color, albedo, and roughness estimation, demonstrating improved material decomposition across diverse scenes. Moreover, our method significantly outperforms all other approaches in relighting quality. This consistent improvement suggests that the additional flashlight information helps our method better disentangle material, geometry, and illumination, thereby reducing the inherent ambiguities that often degrade relighting quality. Our method also achieves competitive runtime efficiency. 

To evaluate the robustness of our method to imperfect segmentation, we additionally replace the ground-truth segmentation masks with per-view masks generated by SAM2~\cite{ravi2025sam}, denoted as ``Ours-SAM2''. Despite using generated segmentation masks, Ours-SAM2 still outperforms the other methods in diffuse color, albedo, and relighting. Its roughness estimation is slightly worse than that of IRIS, although IRIS is provided with ground-truth segmentation masks. These results demonstrate that our method remains effective without relying on ground-truth 2D segmentation.

Fig.~\ref{fig:comparison-1-1} and~\ref{fig:comparison-1-2} present qualitative comparisons of material decomposition and relighting on representative synthetic scenes. Our method produces more accurate and consistent albedo maps, faithfully capturing the base colors of objects without contamination from shading or specular highlights, as illustrated by the right cabinet door in the kitchen scene and the display cabinet in the market scene. It also achieves the most accurate relighting results, with specular highlights that closely match the ground truth in both spatial distribution and intensity.

For WildLight, which is designed primarily for single-object scenes with relatively dense view sampling, the sparse view sampling in our setting results in less accurate albedo and roughness estimation. For FIPT and IRIS, the estimated roughness is also less accurate than that reported in their original papers. This difference may be attributed to the input-view coverage: our method primarily captures only a portion of the scene, and we supplement the captured images with a limited number of additional views to provide sufficient coverage of the full scene for comparison.

\paragraph{Real-world scenes}

For real-world captures, ground-truth material annotations are unavailable. We therefore mainly provide qualitative comparisons of material decomposition and relighting (Fig.~\ref{fig:comparison-2}). Specifically, we visualize rendered RGB images, diffuse color or albedo, roughness, normal maps, and relighting results when supported by each method. These comparisons are used to evaluate whether each method can recover physically plausible materials and generalize to novel flash illumination. As shown in Fig.~\ref{fig:comparison-2}, for the Coffee Bar scene, our method successfully recovers diffuse materials, such as the fabric of the armchair, as well as specular components, including the chair legs, floor, and table. This improvement is particularly evident in the roughness map, where our results exhibit more accurate and coherent material properties. For the Classroom scene, our method also successfully estimates the specular properties of the whiteboard. Moreover, by leveraging flash images, our method more faithfully recovers the albedo of the wooden podium floor, whereas MVInverse suffers from noticeable color shifts in these regions. For the Car scene, our method also produces plausible albedo and roughness estimates. Notably, our method accurately identifies the black region on the left side of the car window as plastic, which is clearly reflected in the roughness map. As a result, for real-world scenes, our method achieves high-quality material estimation, enabling effective relighting, as shown in the rightmost column.

\begin{table}[t]
\centering
\small
\caption{Quantitative ablation on synthetic scenes, averaged over all scenes. We report PSNR for albedo and diffuse color, and MAE for roughness and surface normals.}
\label{tab:ablation_quantitative}
\setlength{\tabcolsep}{0.5pt}
\begin{tabular}{lcccc}
\toprule
\multirow{2}{*}{Models} 
& Albedo $\uparrow$ 
& Diffuse Color $\uparrow$ 
& Roughness 
& Normal \\
\cmidrule(lr){2-2}
\cmidrule(lr){3-3}
\cmidrule(lr){4-4}
\cmidrule(lr){5-5}
& PSNR $\uparrow$
& PSNR $\uparrow$
& MAE $\downarrow$ 
& MAE $\downarrow$ \\
\midrule
w/o flash 
& 22.23 & \textbf{27.56} & 0.2568 & 0.0219 \\

w/o GADF 
& 23.97 & 15.64 & 0.1799 & \textbf{0.0207} \\

w/o MultiEnvMap 
& 24.00 & 27.55 & 0.1647 & 0.0212 \\

Albedo\&roughness NGP
& 23.36 & 27.26 & 0.2829 & 0.0213 \\

Cam\&flash baseline 
& 23.69 & 27.13 & 0.1683 & 0.0216 \\

\midrule
w/o $\mathcal{L}_\mathrm{n}$ 
& 21.71 & 24.40 & 0.2563 & 0.0803 \\

w/o $\mathcal{L}_\mathrm{a}$ 
& 22.29 & 27.38 & 0.1653 & 0.0212 \\

w/o $\mathcal{L}_\mathrm{sem}$ 
& 23.96 & 27.36 & 0.1600 & 0.0227 \\

w/o $\mathcal{L}_\mathrm{s}$ 
& 23.97 & 27.51 & 0.2639 & 0.0212 \\

\midrule
Full 
& \textbf{24.14} & 27.49 & \textbf{0.1554} & 0.0211 \\
\bottomrule
\end{tabular}
\end{table}

\subsection{Ablations}
\label{sec:ablations}

\begin{figure*}[t!]
\centering
\includegraphics[width=0.95\linewidth]{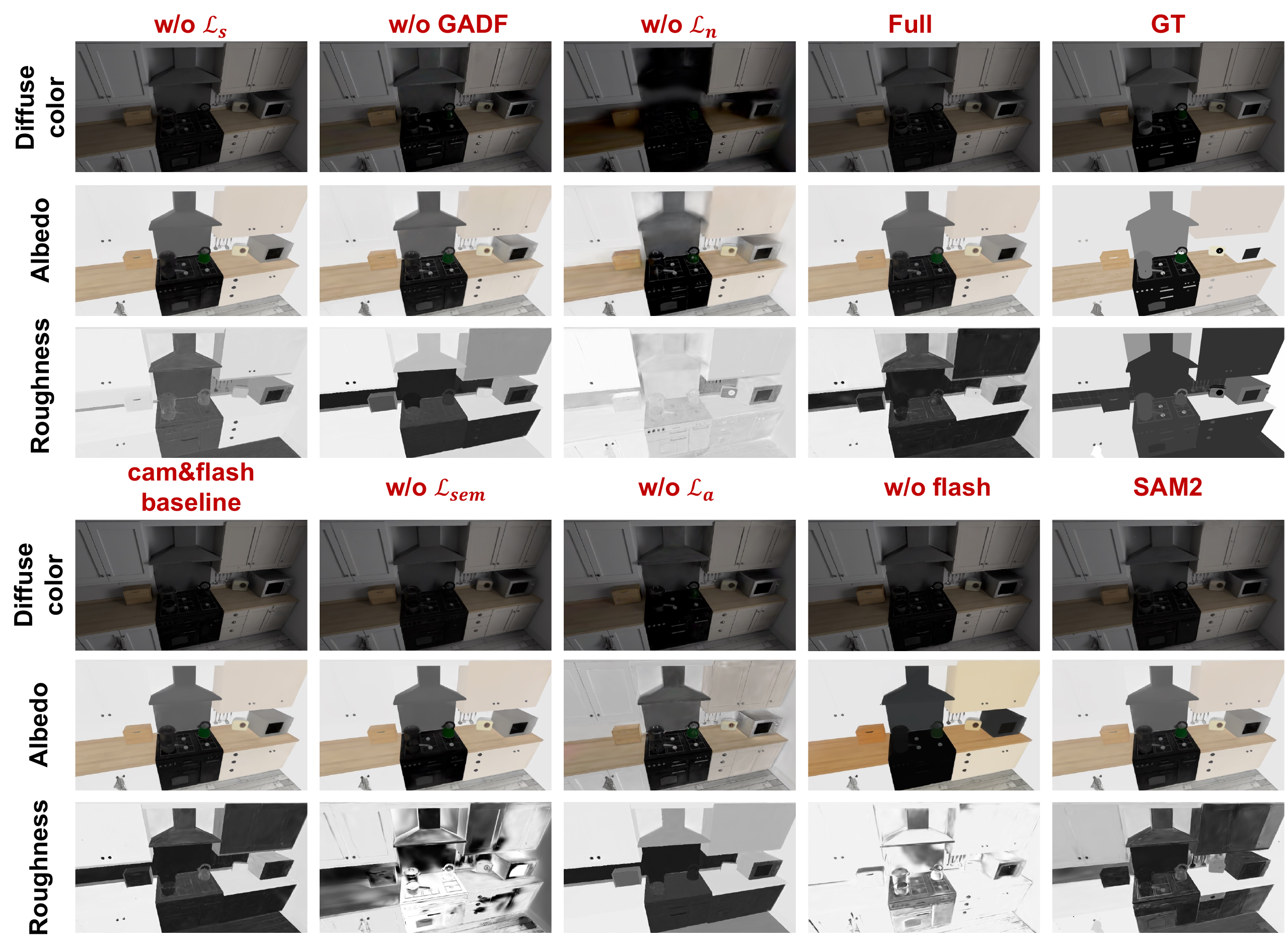}
\caption{\textbf{Qualitative comparison of the ablation studies.}}
\label{fig:ablation-visual-compare} 
\end{figure*}

We conduct ablation studies to validate the contribution of each component of our method (Table~\ref{tab:ablation_quantitative}). Each variant is obtained by removing or replacing one component of the full model while keeping all other settings unchanged. As shown in Table~\ref{tab:ablation_quantitative}, in the first part of the ablation study, we evaluate different configurations, including the presence or absence of flashlight input and GADF. The results demonstrate that flashlight images significantly improve albedo and roughness estimation. Removing the flashlight input slightly improves diffuse-color PSNR by 0.07 dB. This may be because the flashlight primarily provides additional cues for material estimation rather than diffuse-color reconstruction.

Using multiple environment maps provides an improvement in material estimation. We also introduce an offset between the flashlight and the camera, referred to as the ``cam\&flash baseline'', to simulate real-world acquisition conditions. We set the offset to $0.5$, approximately corresponding to the width of a camera in the synthetic scene. The results show that our method still produces robust inverse rendering results even with this relatively large offset.

We further replace the explicit albedo and roughness fields with NGP-based fields, following the same representation used for the GS-anchored diffuse field, while keeping the GS-anchored diffuse field unchanged. We denote this variant as ``Albedo\&roughness NGP''. As shown in Table~\ref{tab:ablation_quantitative}, this variant decreases the albedo PSNR and leads to worse roughness estimation. We speculate that the fine-scale details represented by the NGP fields are less effectively optimized than those explicitly represented by the 2GDS in our original formulation.

Next, we perform an ablation study on the individual loss terms. Removing the normal prior loss decreases the quality of the predicted normals, which in turn affects material estimation. Removing the albedo loss degrades albedo estimation but has relatively little impact on roughness. The specular decomposition loss substantially improves roughness estimation, while the semantic loss provides a smaller but consistent improvement. In Fig.~\ref{fig:ablation-visual-compare}, we present qualitative comparisons of the ablation variants on the Kitchen scene. The visual results are consistent with the quantitative results in Table~\ref{tab:ablation_quantitative}, further demonstrating the effectiveness of our key components and loss terms.

%% file: sec/5_conclusion.tex
\section{Limitations and Future Work}
One limitation of our method is that, compared with one-light-at-a-time methods, it is not sufficiently sensitive to subtle changes in roughness. Moreover, in regions that are overexposed even without flash illumination, our method may lose constraints and produce noisy material estimates. In future work, we plan to explore the use of RAW image formats and camera response function (CRF) modeling to better handle overexposed regions.

\section{Conclusion}
We introduced a practical inverse rendering framework for casually captured indoor scenes, where material and geometry estimation are challenging due to complex scene structure, diverse reflectance properties, occlusions, and indirect light transport. Our key insight is to combine flash and non-flash images as complementary physical cues for more reliable material decomposition.
Based on this insight, our method integrates physically based static and flash rendering with a view-consistent diffuse representation. Across synthetic and real-world scenes, our approach improves material decomposition and relighting quality over recent baselines.